\documentclass[10pt,twocolumn,letterpaper]{article}

\usepackage{cvpr}              

\usepackage[dvipsnames]{xcolor}

\usepackage{graphicx}
\usepackage{amsmath}
\usepackage{amssymb}
\usepackage{booktabs}

\usepackage{amsmath,amsfonts,bm}

\def\eqref#1{equation~\ref{#1}}

\def\1{\bm{1}}

\DeclareMathAlphabet{\mathsfit}{\encodingdefault}{\sfdefault}{m}{sl}
\SetMathAlphabet{\mathsfit}{bold}{\encodingdefault}{\sfdefault}{bx}{n}

\def\sR{{\mathbb{R}}}

\usepackage{algorithm}
\usepackage{algpseudocode}
\usepackage{times}
\usepackage{epsfig}
\usepackage{epigraph}
\usepackage{graphicx}
\usepackage{amsmath}
\usepackage{amssymb}
\usepackage{tabularx}
\usepackage{multirow}
\usepackage{float}
\usepackage{soul}
\usepackage{pifont}
\usepackage{wrapfig}
\usepackage{threeparttable}

\definecolor{cvprblue}{rgb}{0.21,0.49,0.74}
\usepackage[pagebackref,breaklinks,colorlinks,citecolor=cvprbl ue]{hyperref}
\usepackage[accsupp]{axessibility}
\newcommand{\ldist}[0]{\mathcal{L}_{\text{SDS}}}
\newcommand{\x}[0]{\mathbf{x}}
\newcommand{\z}[0]{\mathbf{z}}

\newcommand{\zt}[0]{{\z}_t}

\def\name{GaussianDreamer\xspace}

\def\paperID{2632} 
\def\confName{CVPR}
\def\confYear{2024}

\title{GaussianDreamer: Fast Generation from Text to 3D Gaussians\\by Bridging 2D and 3D Diffusion Models}

\author{
Taoran Yi$^{1}$, \; Jiemin Fang$^{2}$\footnotemark[2],\; Junjie Wang$^{2}$,\; Guanjun Wu$^{3}$, \; Lingxi Xie$^{2}$,\\ Xiaopeng Zhang$^{2}$, \; Wenyu Liu$^1$, \; Qi Tian$^{2}$,\; Xinggang Wang$^1$\footnotemark[2]\ \footnotemark[3]\\
$^1$School of EIC, Huazhong University of Science and Technology \;\;
$^2$Huawei Inc.\\
$^3$School of CS, Huazhong University of Science and Technology\\
\texttt{\small\{taoranyi, guajuwu, liuwy, xgwang\}@hust.edu.cn}\\
\texttt{\small\{jaminfong, is.wangjunjie, 198808xc, zxphistory\}@gmail.com}  \;\;
\texttt{\small tian.qi1@huawei.com}
}

\begin{document}

\twocolumn[{%
\renewcommand\twocolumn[1][]{#1}%
\maketitle
\vspace{-35pt}
\begin{center}
\centering
\includegraphics[width=1.0\linewidth]{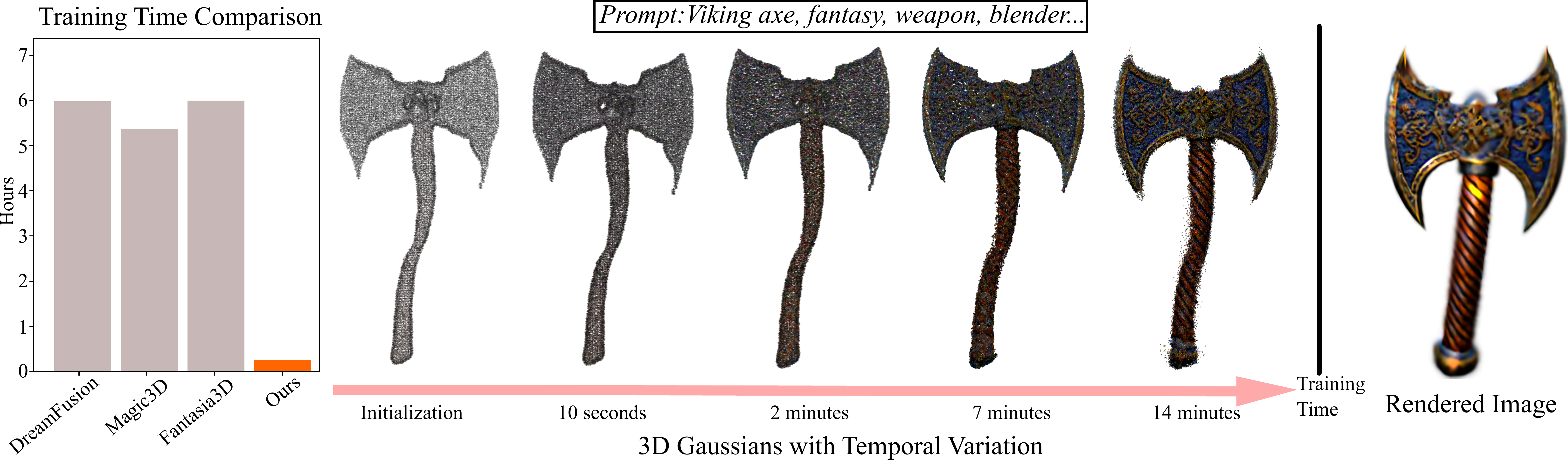}
\vspace{-20pt}
\captionof{figure}{We propose a simple yet efficient framework called GaussianDreamer. It bridges the 3D and 2D diffusion models via Gaussian splatting, having both 3D consistency and rich generation details. Our method can complete training within 15 minutes on a single GPU and achieve real-time rendering.
}
\label{fig: teaser}
\end{center}%
\vspace{-8pt}
}]

{
\renewcommand{\thefootnote}{\fnsymbol{footnote}}
\footnotetext[2]{Project lead.}
\footnotetext[3]{Corresponding author.}
}
\begin{abstract}
In recent times, the generation of 3D assets from text prompts has shown impressive results. Both 2D and 3D diffusion models can help generate decent 3D objects based on prompts. 3D diffusion models have good 3D consistency, but their quality and generalization are limited as trainable 3D data is expensive and hard to obtain. 2D diffusion models enjoy strong abilities of generalization and fine generation, but 3D consistency is hard to guarantee. This paper attempts to bridge the power from the two types of diffusion models via the recent explicit and efficient 3D Gaussian splatting representation. A fast 3D object generation framework, named as GaussianDreamer, is proposed, where the 3D diffusion model provides priors for initialization and the 2D diffusion model enriches the geometry and appearance. Operations of noisy point growing and color perturbation are introduced to enhance the initialized Gaussians. Our GaussianDreamer can generate a high-quality 3D instance or 3D avatar within 15 minutes on one GPU, much faster than previous methods, while the generated instances can be directly rendered in real time. Demos and code are available at \url{https://taoranyi.com/gaussiandreamer/}.


\end{abstract}

\section{Introduction}
\label{sec: Introduction}
3D asset generation has been an expensive and professional work in conventional pipelines. Recently, diffusion models~\cite{rombach2022high} have achieved great success in creating high-quality and realistic 2D images. Many research works~\cite{tang2023dreamgaussian,chen2023text,poole2022dreamfusion,lin2023magic3d,wang2023score,chen2023fantasia3d,wang2023prolificdreamer,xu2023dream3d,shi2023mvdream,zhao2023efficientdreamer,armandpour2023re,jun2023shap,nichol2022point,gupta3dgen,gao2022get3d} try to transfer the power of 2D diffusion models to the 3D field for easing and assisting the process of 3D assets creation, \eg the most common text-to-3D task. 

Here come two main streams for achieving this goal: (i) training a new diffusion model with 3D data~\cite{jun2023shap,nichol2022point,gupta3dgen,gao2022get3d} (namely the \textit{3D diffusion model}) and (ii) lifting the \textit{2D diffusion model} to 3D~\cite{tang2023dreamgaussian,chen2023text,poole2022dreamfusion,lin2023magic3d,wang2023score,chen2023fantasia3d,wang2023prolificdreamer,xu2023dream3d,shi2023mvdream,zhao2023efficientdreamer,armandpour2023re,rombach2022high}. The former one is direct to implement and holds strong 3D consistency, but struggles to extend into a large generation domain as 3D data is usually hard and expensive to obtain. The scale of current 3D datasets is far smaller than 2D datasets. This results in the generated 3D assets falling short in dealing with complex text prompts and producing complex/fine geometry and appearance. The latter benefits from the large data domain of the 2D diffusion models, which can handle various text prompts and produce highly detailed and complex geometry and appearance. However, as 2D diffusion models are unaware of the camera view, the generated 3D assets are hard to form geometry consistency, especially for structure-complicated instances. 

This paper proposes to use recent 3D Gaussian Splatting~\cite{kerbl3Dgaussians} to bridge the two aforementioned approaches, simultaneously having the geometry consistency from 3D diffusion models and rich details from 2D diffusion models. 3D Gaussians are one type of efficient and explicit representation, which intrinsically enjoys geometry priors due to the point-cloud-like structure. 
Specifically, we use one of two types of 3D diffusion models: text-to-3D and text-to-motion diffusion models, \eg Shap-E~\cite{jun2023shap} and MDM~\cite{tevet2022mdm} in our implementation, to generate a coarse 3D instance. Based on the coarse 3D instance, a group of 3D Gaussians are initialized. We introduce two operations of \textbf{noisy point growing} and \textbf{color perturbation} to supplement the initialized Gaussians for follow-up enriching the 3D instance. Then the 3D Gaussians can be improved and optimized by interacting with the 2D diffusion model via the Score Distillation Sampling~\cite{poole2022dreamfusion} (SDS) loss. Due to the geometry priors from both the 3D diffusion model and 3D Gaussian Splatting itself, the training process can be finished in a very short time. The generated 3D asset can be rendered in real time without transformation into structures like mesh via the splatting process.

Our contributions can be summarized as follows.
\begin{itemize}
    \item We propose a text-to-3D method, named as \textbf{\name} which bridges the 3D and 2D diffusion models via Gaussian splitting, enjoying both 3D consistency and rich generation details.
    \item Noisy point growing and color perturbation are introduced to supplement the initialized 3D Gaussians for further content enrichment.
    \item The overall method is simple and quite effective. A 3D instance can be generated within 15 minutes on one GPU, much faster than previous methods, and can be directly rendered in real time.
\end{itemize}

\begin{figure*}[thbp]
    \centering
    \includegraphics[width=\linewidth]{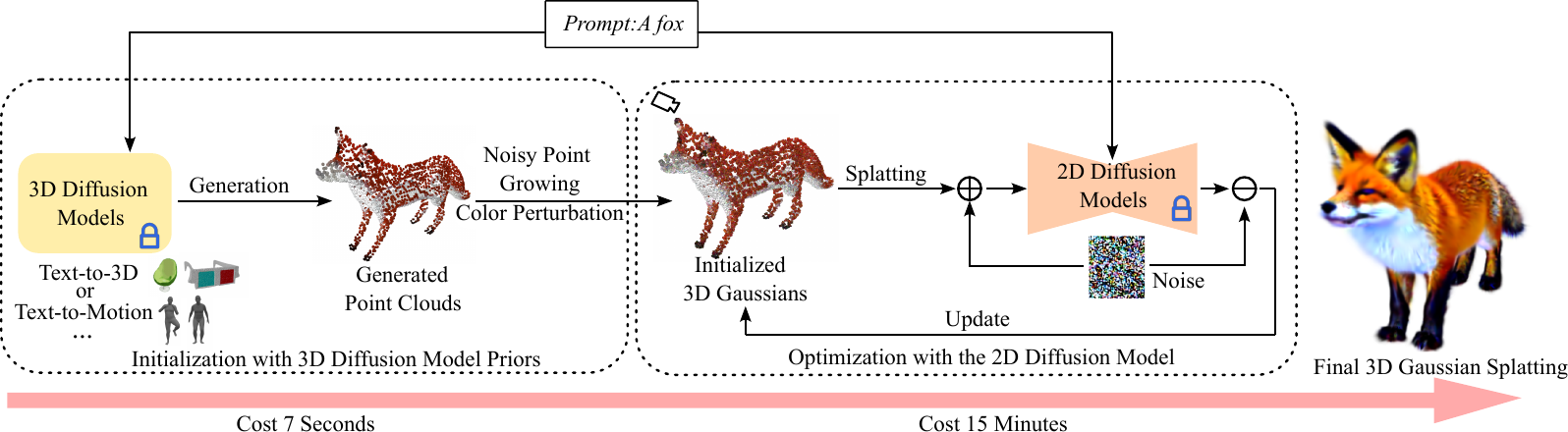}
    \vspace{-15pt}
    \caption{Overall framework of GaussianDreamer. Firstly, we utilize a 3D diffusion model to generate the initialized point clouds. After executing noisy point growing and color perturbation on the point clouds, we use them to initialize the 3D Gaussians. The initialized 3D Gaussians are further optimized using the SDS method~\cite{poole2022dreamfusion} with a 2D diffusion model. Finally, we render the image using the 3D Gaussians by employing 3D Gaussian Splatting~\cite{kerbl3Dgaussians}. We can use one of various 3D diffusion models to generate the initialized point clouds. In this case, we take text-to-3D and text-to-motion diffusion models as examples.}
    \label{fig: framework}
    \vspace{-15pt}
    \end{figure*}
    
\section{Related Works}
\label{sec: related}

\paragraph{3D Pretrained Diffusion Models.}
Recently, text-to-3D asset generation using diffusion models has achieved great success. Currently, it is mainly divided into lifting 2D diffusion models to 3D and 3D pretrained diffusion models, the difference lies in whether the training data used is 2D or 3D.
3D pretrained diffusion models~\cite{jun2023shap,nichol2022point,gao2022get3d,gupta3dgen}, referred to as 3D diffusion models in our paper, are models pretrained on text-3D pairs. After pretraining, they can generate 3D assets through only inference, and models such as Point-E~\cite{nichol2022point} and Shape-E~\cite{jun2023shap} can generate 3D assets in minutes.
In addition to generating 3D assets from text, there are methods where 3D diffusion models~\cite{tevet2022mdm, zhao2023modiff,raab2023single,zhang2023remodiffuse,chen2023executing,kim2023flame,zhou2023ude,dabral2023mofusion,ao2023gesturediffuclip,shafir2023human} generate motion sequences based on text-motion data. By pretraining on text-motion pairs, these models can generate reasonable motion sequences for different texts. The generated motion sequences can be transformed into the SMPL (Skinned Multi-Person Linear) model~\cite{loper2023smpl} based on mesh representation, but texture information is not included. In our method, we can paint transformed SMPL by using different text prompts.

\paragraph{Lifting 2D Diffusion Models to 3D.}
In text-to-3D asset generation methods~\cite{chen2023it3d,seo2023ditto,seo2023let,tsalicoglou2023textmesh,wu2023hd,yu2023points,zhu2023hifa,raj2023dreambooth3d,qi2023vpp,lorraine2023att3d,huang2023tech,ouyang2023chasing,park2023ed,jiang2023avatarcraft,zhang2023dreamface,han2023headsculpt,song2023roomdreamer}, in addition to using 3D pretrained diffusion models, lifting a 2D diffusion model to 3D is a training-free approach. Moreover, due to the abundance of 2D image data, this method produces assets with higher diversity and fidelity. Some single-image-to-3D methods~\cite{tang2023make,liu2023one,liu2023zero1to3,qian2023magic123,shi2023zero123++,lin2023consistent123,shi2023toss,sargent2023zeronvs,liu2023syncdreamer,long2023wonder3d,ye2023consistent,weng2023zeroavatar,yang2023consistnet,weng2023consistent123,hu2023humanliff,gu2023nerfdiff,li20223ddesigner} also employ similar ideas.
DreamFusion~\cite{poole2022dreamfusion} first proposes SDS (Score Distillation Sampling) method, which is to update the 3D representation model using the 2D diffusion model. 
\cite{wang2023score} proposes the method of SJC (Score Jacobian Chaining) to lift the 2D diffusion model to 3D. Later methods~\cite{lin2023magic3d,chen2023fantasia3d,wang2023prolificdreamer,li2023sweetdreamer,sun2023dreamcraft3d} build on DreamFusion and further improve the quality of 3D generation. Among them, generated 3D assets may suffer from multi-face problems. To address this issue, some methods strengthen the semantics of different views~\cite{armandpour2023re} and use multi-view information~\cite{zhao2023efficientdreamer,shi2023mvdream} to alleviate such problems. 
There are also models ~\cite{sanghi2022clip,jain2022zero,michel2022text2mesh,lei2022tango,mohammad2022clip,wang2022clip,xu2023dream3d,hong2022avatarclip} which adopt CLIP~\cite{radford2021learning} to align each view of the 3D representation model with the text.

\paragraph{3D Representation Methods.}

In recent times, neural radiance fields (NeRF)~\cite{mildenhall2020nerf} have achieved impressive results in 3D representation, and many methods in text-to-3D asset generation have also adopted NeRF or its variants~\cite{barron2021mip,mueller2022instant} as the representation method. 
Some methods~\cite{chen2023fantasia3d,lin2023magic3d,li2023sweetdreamer,gao2022get3d} use explicit optimizable mesh representation methods like DMTET~\cite{shen2021deep} to reduce rendering costs and further improve resolution. 
In addition to that, there are also generation methods that utilize point clouds~\cite{nichol2022point,luo2021diffusion,vahdat2022lion,qi2023vpp} and meshes~\cite{liu2023meshdiffusion} as 3D representations.

Recently, 3D Gaussian Splatting~\cite{kerbl3Dgaussians} has been introduced as a representation method for 3D scenes, which can achieve rendering effects comparable to NeRF-based methods and enable real-time rendering. 
Two concurrent works~\cite{tang2023dreamgaussian,chen2023text} also construct the 3D representation with 3D Gaussian Splatting~\cite{kerbl3Dgaussians}. DreamGaussian~\cite{tang2023dreamgaussian} uses a single image as a condition to generate 3D assets, while GSGEN~\cite{chen2023text} implements high-quality generation from text to 3D. 
Our method shares a similar idea of using 3D Gaussian Splatting as the representation method, which significantly reduces the cost of improving resolution and achieving a much faster optimization speed compared to optimizable mesh representation methods. And we can generate detailed 3D assets based on prompt texts in a very short time.

\vspace{-5pt}
\section{Method}
In this section, we first review 2D and 3D diffusion models and the 3D representation method -- 3D Gaussian Splatting~\cite{kerbl3Dgaussians}. We give an overview of the whole framework in Sec.~\ref{subsec: framework}. Then, in Sec.~\ref{subsec: initialization}, we describe the process of initializing the 3D Gaussians with the assistance of 3D diffusion models. The further optimization of 3D Gaussians using the 2D diffusion model is described in Sec.~\ref{subsec: Optimize}.

\subsection{Preliminaries}
\label{subsec: pre}
\paragraph{DreamFusion.}
DreamFusion~\cite{poole2022dreamfusion} is one of the most representative methods to lift 2D diffusion models to 3D, which proposes to optimize the 3D representation with the score distillation sampling (SDS) loss via a pre-trained 2D diffusion model $\phi$. Specifically, it takes MipNeRF~\cite{barron2021mip} as the 3D representation method, whose parameters $\theta$ are optimized. Taking the rendering method as $g$, the rendered image results in $\x = g(\theta)$. To make the rendered image $\x$ similar to the samples obtained from the diffusion model $\phi$, DreamFusion uses a scoring estimation function: $\hat\epsilon_\phi(\zt; y, t)$, which predicts the sampled noise $\hat\epsilon_\phi$ given the noisy image $\zt$, text embedding $y$, and noise level $t$.
By measuring the difference between the Gaussian noise $\epsilon$ added to the rendered image $\x$ and the predicted noise $\hat\epsilon_\phi$, this scoring estimation function can provide the direction for updating the parameter $\theta$. 
The formula for computing the gradient is as
\begin{align}
    \nabla_{\theta} \ldist(\phi, \x=g(\theta)) \triangleq \mathbb{E}_{t, \epsilon}\left[w(t)\left(\hat\epsilon_\phi(\zt; y, t)  - \epsilon\right) {\partial \x \over \partial \theta}\right],
    \label{eq: sdsgrad}
\end{align}
where $w(t)$ is a weighting function.

\vspace{-5pt}
\paragraph{3D Gaussian Splatting.}
3D Gaussian Splatting~\cite{kerbl3Dgaussians} (3D-GS) is a recent groundbreaking method for novel-view synthesis. Unlike implicit representation methods such as NeRF~\cite{mildenhall2020nerf}, which renders images based on volume rendering, 3D-GS renders images through splatting~\cite{yifan2019differentiable}, achieving real-time speed. Specifically, 3D-GS represents the scene through a set of anisotropic Gaussians, defined with its center position $\mu \in \sR^3$, covariance $\Sigma \in \sR^7$, color $c \in \sR^3$, and opacity $\alpha \in \sR^1$.
And the 3D Gaussians can be queried as follows:
\begin{equation}
	\label{eq: gaussian}
	G(x)~= e^{-\frac{1}{2}(x)^{T}\Sigma^{-1}(x)},
\end{equation}
where $x$ represents the distance between $\mu$ and the query point.
For computing the color of each pixel, it uses a typical neural point-based rendering~\cite{kopanas2022neural,kopanas2021point}. A ray $r$ is cast from the center of the camera, and the color and density of the 3D Gaussians that the ray intersects are computed along the ray. The rendering process is as follows:
\begin{equation}
	\label{eq: front-to-back}
	C(r) = \sum_{i \in \mathcal{N}}
	c_{i}\sigma_{i}
	\prod_{j=1}^{i-1}(1-\sigma_{j}), \quad
        \sigma_{i} = \alpha_{i} G(x_i),
\end{equation}
where $\mathcal{N}$ represents the number of sample points on the ray $r$, $c_{i}$ and $\alpha_{i}$ denote the color and opacity of the i-th Gaussian, and $x_i$ is the distance between the point and the i-th Gaussian.


%
\subsection{Overall Framework}
\label{subsec: framework}
Our overall framework consists of two parts, initialization with 3D diffusion model priors and optimization with the 2D diffusion model, as shown in Fig.~\ref{fig: framework}. For initialization with 3D diffusion model priors, we use the 3D diffusion models $F_{3D}$, instantiated with the text-to-3D and text-to-motion diffusion models, to generate the triangle mesh $m$ based on the text prompt $y$, which can be denoted as $m = F_{3D}(y)$. One set of generated point clouds is transformed from the mesh $m$. Then the 3D Gaussians $\theta_b$ are initialized via the generated point clouds after noisy point growing and color perturbation. For better quality, we utilize the 2D diffusion model $F_{2D}$ to further optimize the initialized 3D Gaussians $\theta_b$ via SDS~\cite{poole2022dreamfusion} with prompts $y$, resulting in the final 3D Gaussians $\theta_f$. The target instance can be rendered in real time by splatting the generated Gaussians.

\subsection{Gaussian Initialization with 3D Diffusion Model Priors}
\label{subsec: initialization}


In this section, we mainly discuss how to initialize the 3D Gaussians with 3D diffusion model priors.
First, we use the 3D diffusion model $F_{3D}$ to generate 3D assets based on the prompts $y$. Then we convert the 3D assets into point clouds and use the transformed point clouds to initialize the 3D Gaussians. We have employed two types of 3D diffusion models to generate 3D assets. Below, we explain how to initialize the 3D Gaussians using each of these models.
\vspace{-5pt}
\subsubsection{Text-to-3D Diffusion Model}
\label{subsubsec: Text-to-3D}

When using a text-based 3D generation model, generated 3D assets employ multi-layer perceptrons (MLPs) to predict SDF values and texture colors. To construct the triangle mesh $m$, we query the SDF values at vertices along a regular grid with the size of $128^3$ within the MLPs. Then we query the texture colors at each vertex of $m$. We convert the vertices and colors of $m$ into point clouds, denoted as $\bm{pt}_m(\bm{p}_m,\bm{c}_m)$. $\bm{p}_m \in \sR^3$ refers to the position of the point clouds, which equals to the vertice coordinated of $m$. $\bm{c}_m \in \sR^3$ refers to the color of the point clouds, which is the same as the color of $m$. 
However, the obtained colors $\bm{c}_m$ are relatively simple, and the positions $\bm{p}_m$ are sparse.

\begin{figure}[t]
   \centering
   \includegraphics[width= 0.95 \linewidth]{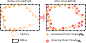}
   \caption{The process of noisy point growing and color perturbation. ``Grow\&Pertb.'' denotes noisy point growing and color perturbation.}
   \label{fig: randompoint}
    \vspace{-12pt}
\end{figure}

\vspace{-5pt}
\paragraph{Noisy Point Growing and Color Perturbation.}
We do not use the generated point clouds $\bm{pt}_m$ to directly initialize the 3D Gaussians. To improve the quality of initialization, we perform noisy point growing and color perturbation around point clouds $\bm{pt}_m$. First, we compute the bounding box (BBox) of the surface on $\bm{pt}_m$ and then uniformly grow point clouds $\bm{pt}_r(\bm{p}_{r}, \bm{c}_{r})$ within the BBox. 
$\bm{p}_{r}$ and $\bm{c}_{r}$ represent the positions and colors of $\bm{pt}_r$. To enable fast searching, we construct a KDTree~\cite{bentley1975multidimensional} $K_m$ using the positions $\bm{p}_m$. Based on the distances between the positions $\bm{p}_{r}$ and the nearest points found in the KDTree $K_m$, we determine which points to keep. In this process, we select points within the (normalized) distance of 0.01. For the noisy point clouds, we make their colors $\bm{c}_{r}$ similar to $\bm{c}_m$, and also add some perturbations: 
\begin{equation}
    \bm{c}_{r} = \bm{c}_m + \bm{a},
\end{equation}
where the values of $\bm{a}$ are randomly sampled between 0 and 0.2. We merge the positions and colors of $\bm{pt}_m$ and $\bm{pt_r}$ to obtain the final point clouds.
\begin{equation}
    \bm{pt}(\bm{p}_f, \bm{c}_f) = (\bm{p}_m \oplus \bm{p}_r,\bm{c}_m \oplus \bm{c}_r),
\end{equation}
where $\oplus$ is the concatenation operation. Fig.~\ref{fig: randompoint} illustrates the process of noisy point growing and color perturbation.
Finally, we initialize the positions $\mu_b$ and colors $c_b$ of the 3D Gaussians $\theta_b(\mu_b,c_b,\Sigma_b, \alpha_b)$ using both the positions $\bm{p}_f$ and colors $\bm{c}_f$ of final point clouds $\bm{pt}$. The opacity $\alpha_b$ of the 3D Gaussians is initialized to 0.1, and the covariance $\Sigma_b$ is calculated as the distance between the nearest two points. Algorithm \ref{alg: add noise} shows the specific algorithm flowchart.

\begin{algorithm}[htbp]
    \caption{The 3D Gaussian Initialization.
    \\ $\bm{pt}_m(\bm{p}_{m}, \bm{c}_{m})$: Point clouds generated from $F_{3D}$.
    \\ $\bm{pt}_r(\bm{p}_{r}, \bm{c}_{r})$: Growing point clouds within the BBox.
    \\ $\bm{pt}(\bm{p}_{f}, \bm{c}_{f})$: Point clouds used for initializing the 3D Gaussians.
    \\ $\theta_b(\mu_b,c_b,\Sigma_b, \alpha_b)$: Initialized 3D Gaussians.}
    \label{alg: add noise}
    \begin{algorithmic}
        \State \textit{Stage 1: Generate points in the BBox.}
        \State $K_m \gets$ BuildKDTree($\bm{p}_{m}$) \Comment{KDTree}
        \State BBox $\gets \bm{p}_{m}$ \Comment{Positions bounding box.}
        \State Low, High $\gets$ BBox.MinBound, BBox.MaxBound \\ \Comment{Boundary of bounding box.}
        \State $\bm{ps}_u$ $\gets$ Uniform(Low, High, size = (NumPoints, $3$))\\ \Comment{Points in the BBox.}
        
        \State \textit{Stage 2: Keep the points that meet the distance requirement.}
        \State $\bm{p}_{r}, \bm{c}_{r}$ = [ ], [ ]
        \ForAll{$\bm{p}_u$ in $\bm{ps}_u$}
            \State  $\bm{p}_{un}, i \gets K_m$.SearchNearest($\bm{p}_u$)\\ \Comment{Nearest point and its index in $K_{m}$.}
            \If{$|\bm{p}_{un} - \bm{p}_u| <0.01 $}
                \State $\bm{p}_{r}$.append($\bm{p}_u$)
                \State $\bm{c}_{r}$.append($\bm{c}_{m}[i]+ 0.2\times$Random(size = 3))
                \\ \Comment{Color of the nearest point plus perturbation.}
            \EndIf
        \EndFor
        \State $\bm{p}_{f} \gets \bm{p}_m \oplus \bm{p}_r$
        \State $\bm{c}_{f} \gets \bm{c}_m \oplus \bm{c}_r$
        \State \textit{Stage 3: Initialize the 3D Gaussians.}
        \State $\mu_b, c_b \gets \bm{p}_{f}, \bm{c}_{f}$ \\  \Comment{Positions and colors of the 3D Gaussians.}
        \State $D \gets \mu_b$ \Comment{Distance between the nearest two positions.}
        \State $\Sigma_b, \alpha_b \gets D, 0.1$	\\ \Comment{Covariance and opacity of the 3D Gaussians.}
    \end{algorithmic}
    \vspace{-4pt}
\end{algorithm}
\vspace{-10pt}

\subsubsection{Text-to-Motion Diffusion Model}

We generate a sequence of human body motions using text and select a human pose that best matches the given text. We then convert the keypoints of this human pose into the SMPL model~\cite{loper2023smpl}, which is represented by a triangle mesh $m$. We then convert the mesh $m$ into point clouds $\bm{pt}_m(\bm{p}_m,\bm{c}_m)$, where the position $\bm{p}_m$ of each point in the point clouds corresponds to the vertices of $m$. As for the color $\bm{c}_m $ of $\bm{pt}_m$, since the SMPL model used here does not have textures, we randomly initialize $\bm{c}_m$.
To move $\bm{pt}_m$ near the origin, we calculate the center point $\bm{p}_c \in \sR^3$ of $\bm{p}_m$ and subtract the position of point clouds $\bm{pt}_m$ from the center point $\bm{p}_c$. 
\begin{equation}
    \bm{pt}(\bm{p}_f, \bm{c}_f) = \bm{pt}_m(\bm{p}_m - \bm{p}_c,\bm{c}_m),
\end{equation}
Finally, we use the point clouds $\bm{pt}$ to initialize the 3D Gaussians, similar to what is described in Sec.~\ref{subsubsec: Text-to-3D}.
To improve the generation of motion sequences, we simplify the text by retaining only the relevant parts related to the motion and add a subject. For example, if the text prompt is ``Iron man kicks with his left leg", we transform it into ``Someone kicks with the left leg" when generating the motion sequences.

\subsection{Optimization with the 2D Diffusion Model}
\label{subsec: Optimize}
To enrich details and improve the quality of the 3D asset, we optimize the 3D Gaussians $\theta_b$ with a 2D diffusion model $F_{2D}$ after initializing them with 3D diffusion model priors. We employ the SDS (Score Distillation Sampling) loss to optimize the 3D Gaussians. First, we use the method of 3D Gaussian Splatting~\cite{kerbl3Dgaussians} to obtain the rendered image $\x = g(\theta_i)$. Here, $g$ represents the splatting rendering method as in Eq.~\ref{eq: front-to-back}. 
Then, we use Eq.~\ref{eq: sdsgrad} to calculate the gradients for updating the Gaussian parameters $\theta_i$ with the 2D diffusion model $F_{2D}$. After a short optimization period using the 2D diffusion model $F_{2D}$, the final generated 3D instance $\theta_f$ achieves high quality and fidelity on top of the 3D consistency provided by the 3D diffusion model $F_{3D}$.

\begin{figure*}[thbp]
    \centering
    \includegraphics[width=\linewidth]{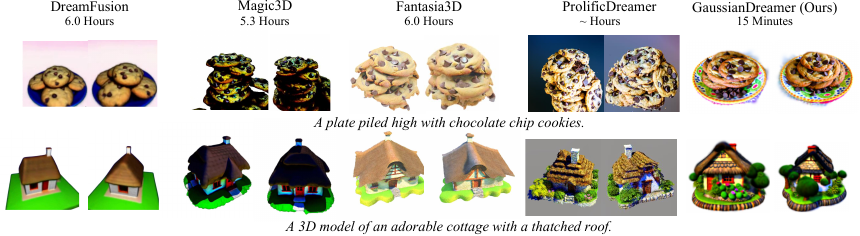}
    \vspace{-18pt}
    \caption{Qualitative comparisons between our method and DreamFusion~\cite{poole2022dreamfusion}, Magic3D~\cite{lin2023magic3d}, Fantasia3D~\cite{chen2023fantasia3d} and ProlificDreamer~\cite{wang2023prolificdreamer}. Here we count the GPU time in their papers. The time for DreamFusion is measured on TPUv4, Magic3D is measured on A100, Fantasia3D is measured on RTX 3090, and our method is measured on RTX 3090.}
    \label{fig: compare3d}
    \vspace{-6pt}
    \end{figure*}
    
\begin{figure*}[thbp]
    \centering
    \includegraphics[width=\linewidth]{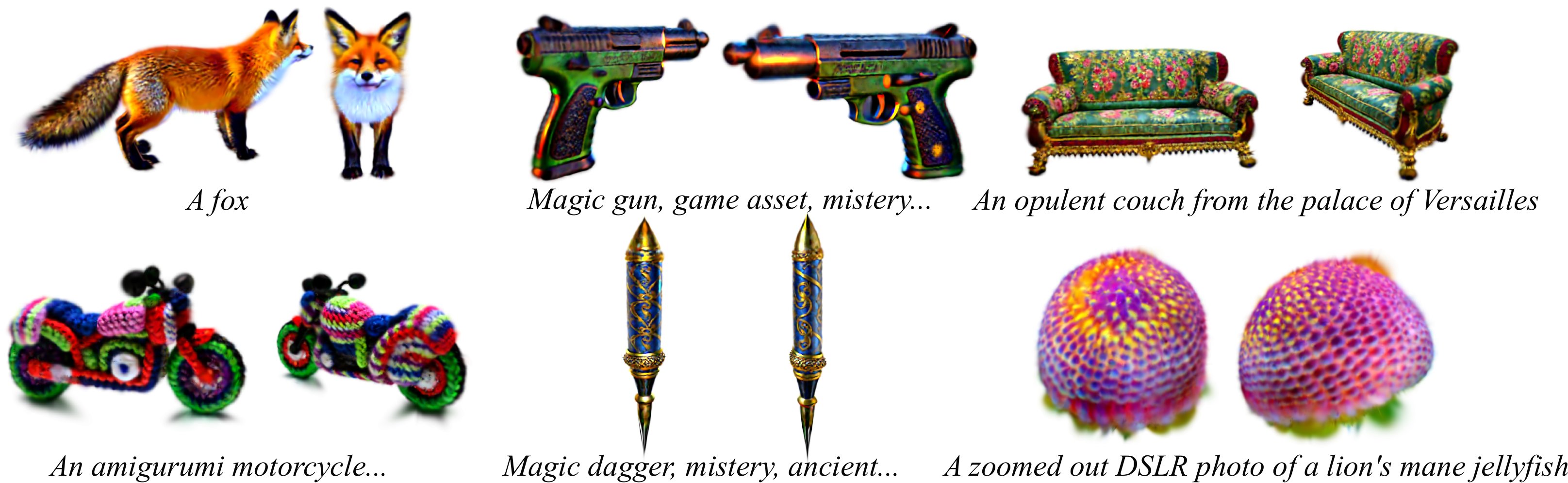}
    \vspace{-18pt}
    \caption{More generated samples by our \name. Two views of each sample are shown.}
    \label{fig: vismore3d}
    \vspace{-15pt}
    \end{figure*}

\begin{figure*}[thbp]
    \centering
    \includegraphics[width=\linewidth]{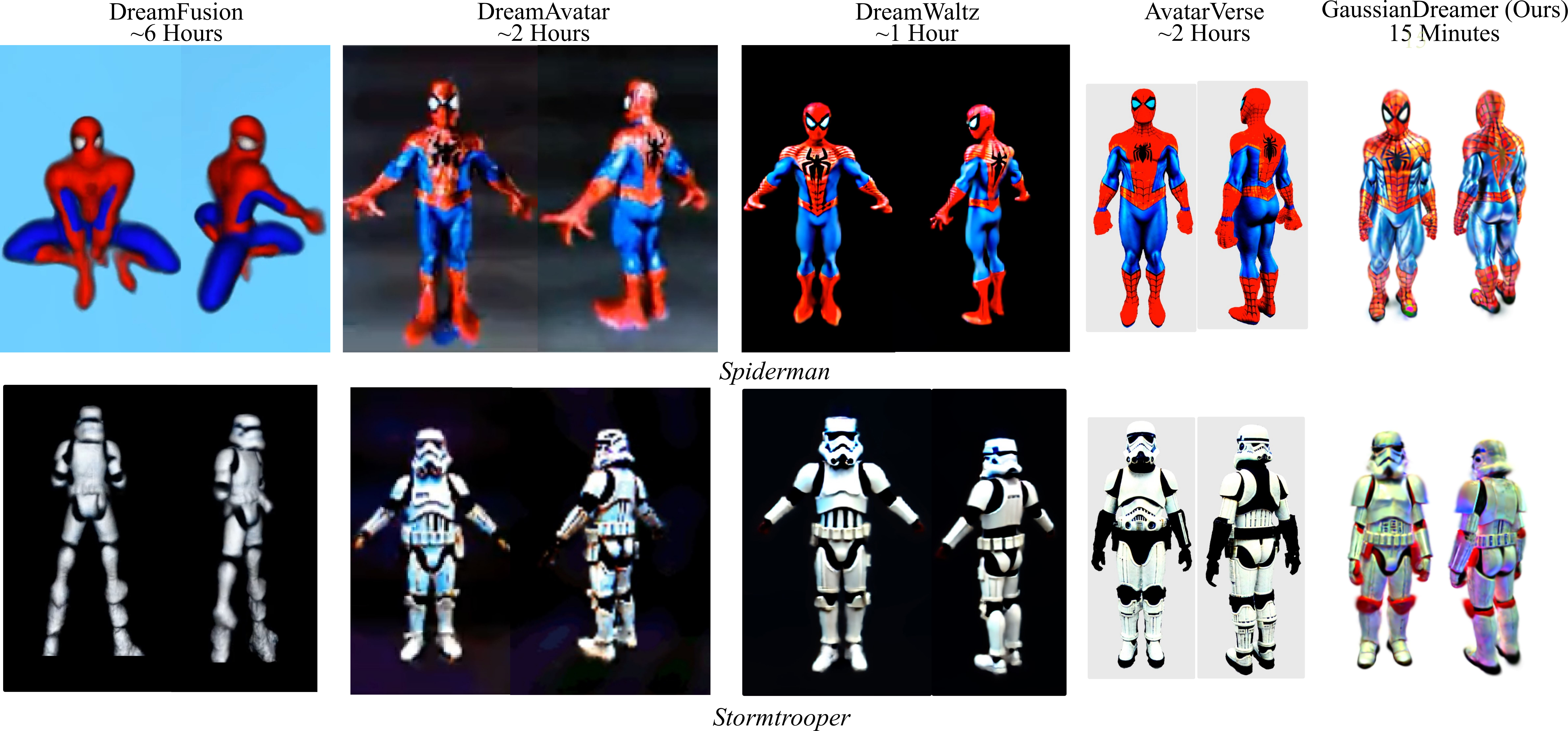}
    \vspace{-20pt}
    \caption{Qualitative comparisons between our method and DreamFusion~\cite{poole2022dreamfusion}, DreamAvatar~\cite{cao2023dreamavatar}, DreamWaltz~\cite{huang2023dreamwaltz}, and AvatarVerse~\cite{zhang2023avatarverse}.}
    \label{fig: comparehuman}
    \vspace{-13pt}
    \end{figure*}

\begin{figure}[t!]
   \centering
   \vspace{-8pt}
   \includegraphics[width= \linewidth]{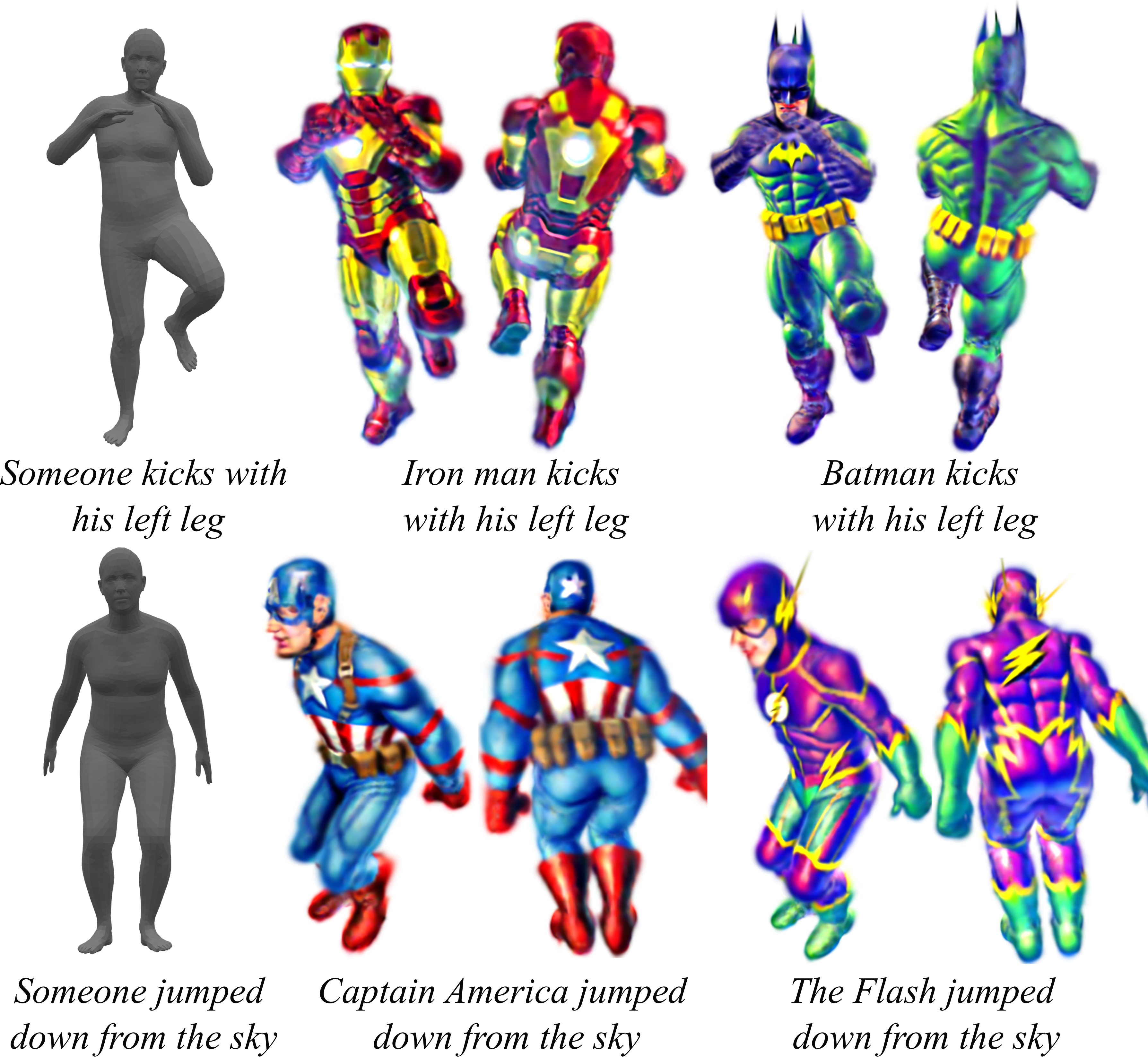}
    \vspace{-20pt}
   \caption{More generated 3D avatars by our GaussianDreamer initialized with the different poses of SMPL~\cite{loper2023smpl}. Here, the different poses of SMPL are generated using a text-to-motion diffusion model.}
   \label{fig: vismorehuman}
    \vspace{-15pt}
\end{figure}

\begin{figure*}[thbp]
    \centering
    \includegraphics[width=\linewidth]{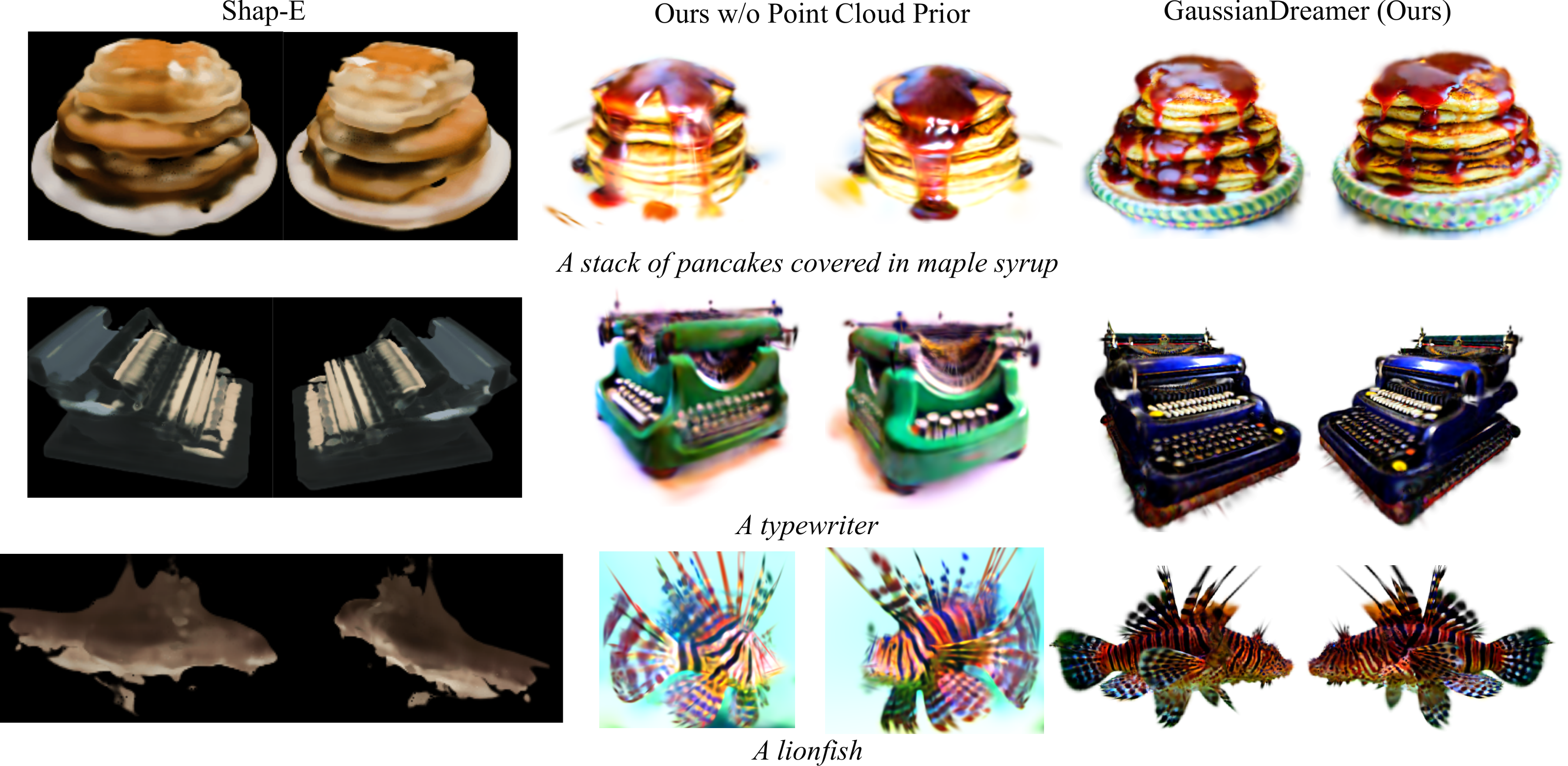}
    \vspace{-10pt}
    \caption{Ablation studies of the initialization of the 3D Gaussians. The Shap-E~\cite{jun2023shap} rendering resolution here is 256x256.}
    \label{fig: abla3d}
    \vspace{-15pt}
    \end{figure*}

\begin{figure}[t!]
   \centering
   \includegraphics[width= \linewidth]{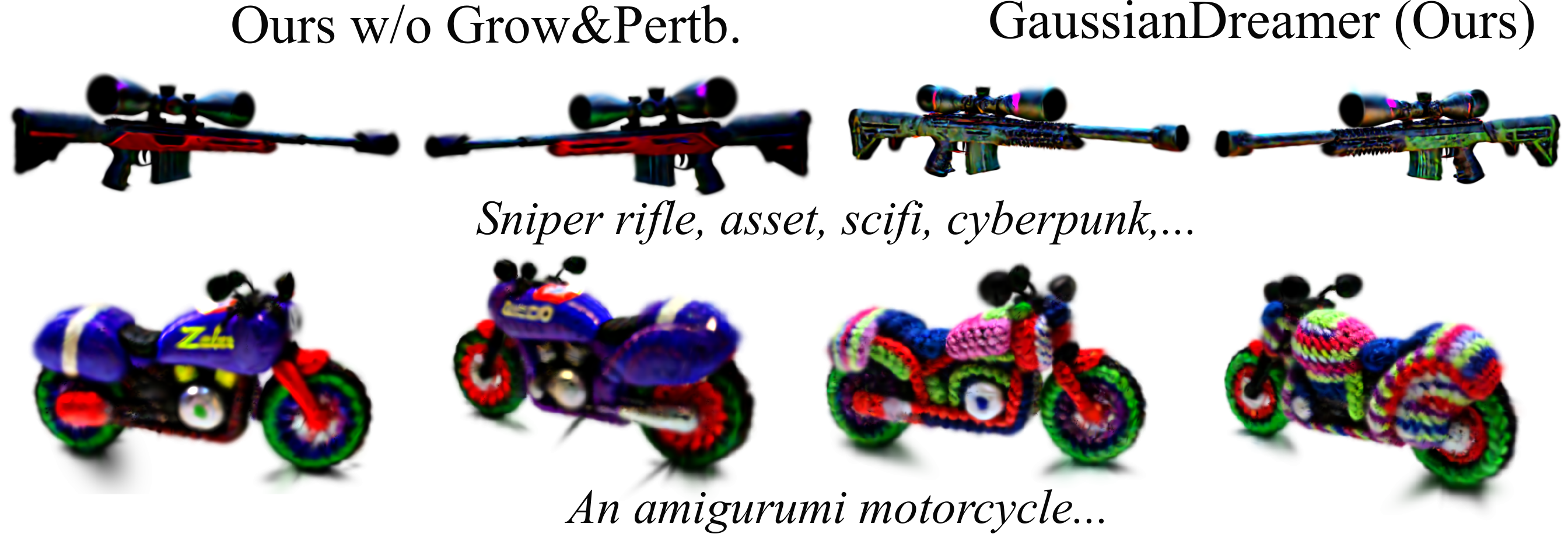}
    \vspace{-20pt}
   \caption{Ablation studies of noisy point growing and color perturbation. ``Grow\&Pertb.'' denotes noisy point growing and color perturbation.}
   \label{fig: ablarandompoint}
    \vspace{-6pt}
\end{figure}

\begin{figure}[t!]
   \centering
   \includegraphics[width= \linewidth]{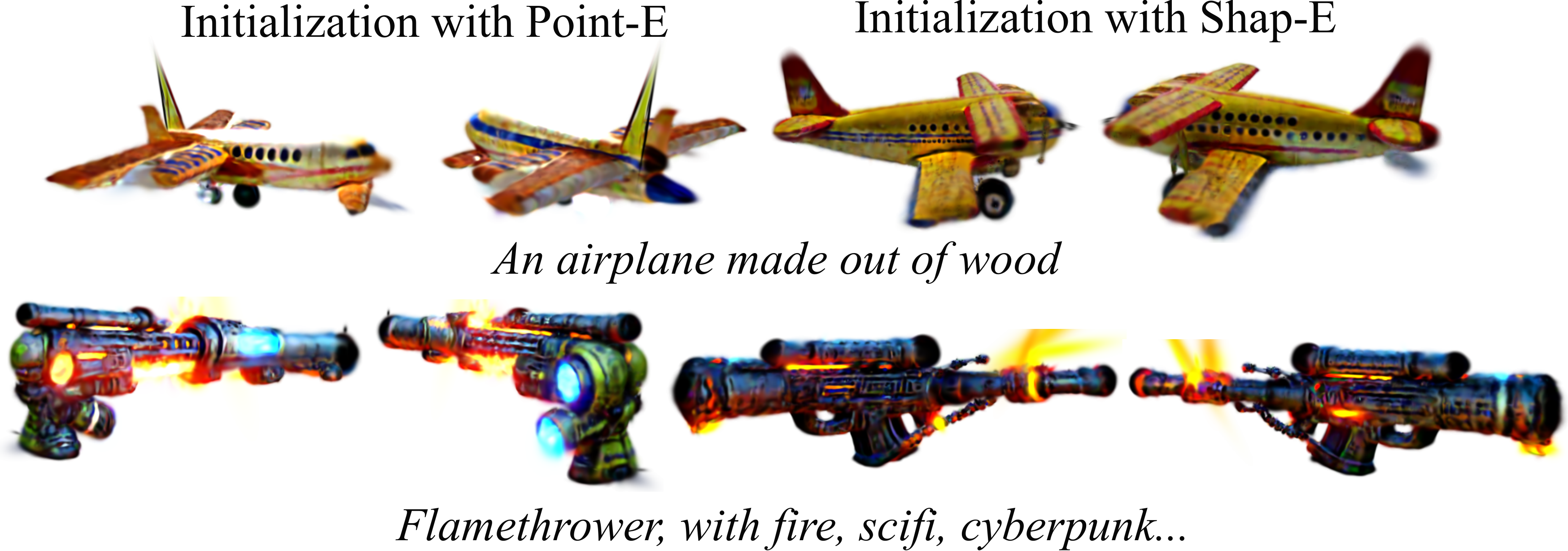}
    \vspace{-20pt}
   \caption{Ablation studies of initialization with different text-to-3D diffusion models: Point-E~\cite{nichol2022point} and Shap-E~\cite{jun2023shap}.}
   \label{fig: initializationdiff}
    \vspace{-12pt}
\end{figure}

\begin{table}[t!]
\setlength{\tabcolsep}{2pt}
\centering
\caption{Quantitative comparisons on T$^3$Bench~\cite{t3bench}.}
\label{tab: t3}
\resizebox{0.5\textwidth}{!}{%
    \begin{tabular}{l|ccccc}
    \toprule
    \textbf{Method}  & \textbf{Time}$^\dag$ & \textbf{Single Obj.} & \textbf{Single w/ Surr.} & \textbf{Multi Obj.} & \textbf{Average}   \\
    \midrule
    SJC~\cite{wang2023score}  & -- &24.7	&19.8&	11.7 &18.7\\
    DreamFusion~\cite{poole2022dreamfusion} & 6 hours & 24.4	&24.6&	16.1& 21.7\\
    Fantasia3D~\cite{chen2023fantasia3d} & 6 hours& 26.4	&27.0&	18.5 &24.0\\
    LatentNeRF~\cite{metzer2023latent}  &  15 minutes &33.1	&30.6&	20.6 &	28.1\\
    Magic3D~\cite{lin2023magic3d}  &5.3 hours &37.0&	35.4&	25.7 &32.7\\
    ProlificDreamer~\cite{wang2023prolificdreamer}	&  $\sim$10 hours &49.4&	44.8&	\textbf{35.8} & 43.3\\
    Ours	 & 15 minutes&\textbf{54.0} &\textbf{48.6}&34.5& \textbf{45.7}\\
    \bottomrule
    \end{tabular}}
\begin{tablenotes}
\footnotesize
\item[]Scores above are average of two metrics (quality and alignment).
\item[]$^\dagger$GPU time counted in their papers.
\end{tablenotes}
\vspace{-10pt}
\end{table}

\section{Experiments}
In this section, we first present the implementation details in Sec.~\ref{subsec: im}. In Sec.~\ref{subsec: quantitative}, we show the quantitative comparison results. Then, in Sec.~\ref{subsec: vis}, we showcase the visualization results of our method and compare them with other methods. In Sec.~\ref{subsec: abla}, we conduct a series of ablation experiments to validate the effectiveness of our method. Finally, we discuss the limitations of our method.
\subsection{Implementation Details}
\label{subsec: im}
Our method is implemented in PyTorch~\cite{paszke2019pytorch}, based on ThreeStudio~\cite{threestudio2023}. The 3D diffusion models used in our method are Shap-E~\cite{jun2023shap} and MDM~\cite{tevet2022mdm}, and we load the Shap-E model finetuned on Objaverse~\cite{deitke2023objaverse} in Cap3D~\cite{luo2023scalable}.
For the 2D diffusion model, we use \emph{stabilityai/stable-diffusion-2-1-base}~\cite{rombach2022high}, with a guidance scale of $100$. The timestamps we use are uniformly sampled from 0.02 to 0.98 before 500 iterations, and change to 0.02 to 0.55 after 500 iterations. 
For the 3D Gaussians, the learning rates of opacity $\alpha$ and position $\mu$ are $10^{-2}$ and $5 \times 10^{-5}$. The color $c$ of the 3d Gaussians is represented by the sh coefficient, with the degree set to 0 and the learning rate set to $1.25 \times 10^{-2}$. The covariance of the 3D Gaussians is converted into scaling and rotation for optimization, with learning rates of $10^{-3}$  and $10^{-2}$, respectively.
The radius of the camera we use for rendering is from 1.5 to 4.0, with the azimuth in the range of -180 to 180 degrees and the elevation in the range of -10 to 60 degrees.
The total training iterations are $1200$. All our experiments can be completed within 15 minutes on a single RTX 3090 with a batch size of 4.
The resolution we use for rendering is 1024 $\times$ 1024, which is scaled to 512 $\times$ 512 when optimizing using the 2D diffusion model. We can render in real time at 512 $\times$ 512 resolution. And all our code will be released.

\subsection{Quantitative Evaluation}
\label{subsec: quantitative}
We evaluate quality and consistency following T$^3$Bench~\cite{t3bench}, which provides a comprehensive benchmark for text-to-3D generation. Three text categories are designed for 3D generation with increasing complexity -- single objects, single objects with surroundings, and multi objects. In Tab.~\ref{tab: t3}, our method outperforms the compared methods while enjoying a short generation time.

\subsection{Visualization Results}
\label{subsec: vis}

In this section, we present the results of initializing the 3D Gaussians with two different 3D diffusion models: text-to-3D and text-to-motion diffusion models.
\vspace{-8pt}
\paragraph{Initialization with Text-to-3D Diffusion Model.}

We show the comparison results with DreamFusion~\cite{poole2022dreamfusion}, Magic3D~\cite{lin2023magic3d}, Fantasia3D~\cite{chen2023fantasia3d} and ProlificDreamer~\cite{wang2023prolificdreamer} in Fig.~\ref{fig: compare3d}. In addition to our method, the figures of other methods are downloaded from the paper of ProlificDreamer.
When encountering prompts that involve the combination of multiple objects, such as the prompt ``A plate piled high with chocolate chip cookies", the generated results from Magic3D, Fantasia3D, and ProlificDreamer do not include a plate. In contrast, our generated result can effectively combine a plate and chocolate chip cookies. Furthermore, compared to DreamFusion, the plate we generate has better patterns.
Our method shows comparable quality while saving $21-24$ times the generation time compared to their methods. Moreover, the 3D Gaussians generated by our method can directly achieve real-time rendering without further transformation into mesh-like structures. Fig.~\ref{fig: vismore3d} visualizes more samples generated by our \name from various prompts, which show good 3D consistency while having high-quality details.
\vspace{-8pt}
\paragraph{Initialization with Text-to-Motion Diffusion Model.}
In Fig.~\ref{fig: comparehuman}, we present the comparison results with DreamFusion~\cite{poole2022dreamfusion}, DreamAvatar~\cite{cao2023dreamavatar}, DreamWaltz~\cite{huang2023dreamwaltz}, and AvatarVerse~\cite{zhang2023avatarverse}. 
In addition to our method, the figures of other methods are downloaded from the paper of AvatarVerse.
It is worth noting that our prompt is ``Spiderman/Stormtrooper stands with open arms", while the prompts for other methods are ``Spiderman/Stormtrooper". This is because when generating motion using text-to-motion diffusion model as initialization, we require more specific action descriptions. Our method achieves a speedup of $4-24$ times compared to other methods, while maintaining comparable quality. Additionally, our method allows for generating 3D avatars with specified body poses. 
In Fig.~\ref{fig: vismorehuman}, we provide more results generated with different human body poses. We first generate a sequence of motions that match the text prompt using a text-to-motion 3d diffusion model, and then initialize the 3d Gaussians with the SMPL of one selected pose in motions.
Our method can generate 3D avatars in any desired pose.
\vspace{-5pt}
\subsection{Ablation Study and Analysis}
\label{subsec: abla}
\paragraph{The Role of Initialization.}
As shown in Fig.~\ref{fig: abla3d}, we first conduct an ablation experiment on the initialization of the 3D Gaussians to validate that initialization can improve 3D consistency. The first column is the rendering result of Shap-E~\cite{jun2023shap} with NeRF~\cite{mildenhall2020nerf} as the 3D representation. The second column is the result of optimizing the 3D Gaussians randomly initialized within a cube using the SDS loss, and the third column is our method. We show the initialization effects on 3 samples. In the first and second rows, Shap-E has good generation results while our method provides more complex geometries and more realistic appearances. Compared with random initialization, in the first row, details of our method are better. In the second row, the 3D assets generated by random initialization have the multi-head problem, which does not occur in our method. The initialization of the 3D diffusion model can avoid unreasonable geometry. In the third row, the generation result of Shap-E is far different from the given text prompt while our method makes the 3D assets closer to the prompt through the 2D diffusion model. Our method can expand the domain of Shap-E prompts, allowing for the generation of 3D assets based on a wider range of prompts.
\vspace{-5pt}
\paragraph{Noisy Point Growing and Color Perturbation.}
Fig.~\ref{fig: ablarandompoint} illustrates the ablation results of noisy point growing and color perturbation. With noisy point growing and color perturbation applied, the first row showcases improved details in the sniper rifle. Additionally, the second column generates an amigurumi motorcycle that better aligns with the style characteristics of amigurumi mentioned in the prompt, compared to the case without noisy point growing and color perturbation.

\vspace{-5pt}
\paragraph{Initialization with Different Text-to-3D Diffusion Models.}
We select two text-to-3D generation models, Shap-E~\cite{jun2023shap} and Point-E~\cite{nichol2022point}, to validate the effectiveness of our framework. 
We load the Point-E model finetuned on Objaverse~\cite{deitke2023objaverse} in Cap3D\cite{luo2023scalable}. Fig.~\ref{fig: initializationdiff}, we showcase the generated results after initializing the 3D Gaussians using one of two text-to-3D generation models. It can be seen that both initializations yield good generation results. However, considering that Shap-E generates 3D assets based on NeRF and SDF, which provide higher fidelity compared to the point cloud representation used by Point-E, the geometry of the airplane in the first row of Fig.~\ref{fig: initializationdiff} appears better when initialized using Shap-E.

\subsection{Limitations}
\label{subsec: lim}
The edges of the 3D assets generated by our method are not always sharp, and there may be unnecessary 3D Gaussians around the object surface. How to filter these point clouds will be a possible direction for improvement.
Our approach utilizes 3D diffusion model priors, which greatly alleviates the issue of multi-face problems. However, there is still a small chance of encountering the multi-face problems in scenes where there is minimal geometric difference but significant appearance discrepancy between the front and back of objects such as a backpack. Utilizing 3D-aware diffusion models~\cite{liu2023zero1to3,shi2023mvdream,zhao2023efficientdreamer} may be able to solve this problem.
Additionally, our method has limited effectiveness in generating large-scale scenes, such as indoor scenes.

\section{Conclusion}
We propose a fast text-to-3D method \name by bridging the abilities of 3D and 2D diffusion models via the Gaussian splatting representation. \name can generate detailed and realistic geometry and appearance while maintaining 3D consistency. The 3D diffusion model priors and geometry priors from the 3D Gaussians effectively promote the convergence speed. Each sample can be generated within 15 minutes on one GPU. We believe the approach of bridging 3D and 2D diffusion models could be a promising direction to generate 3D assets efficiently.

\section*{Acknowledgments}
This work was supported by the National Natural Science Foundation of China (No. 62376102).

{
    \small
    \bibliographystyle{ieeenat_fullname}
    \bibliography{main}
}

\appendix

\section{Appendix}

\begin{figure}[t!]
   \centering
   \includegraphics[width= \linewidth]{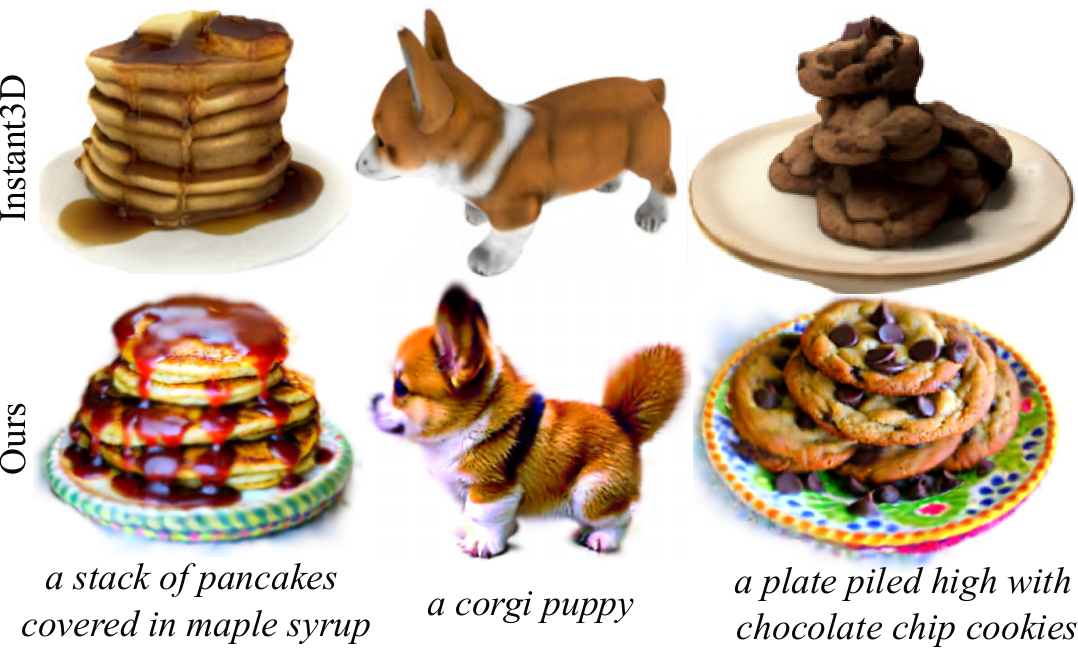}
   \caption{Visual comparisons with Instant3D~\cite{li2023instant3d}.}
   \label{fig: i3d}
\end{figure}

\begin{table*}[t!]
\newcommand{\xmark}{\ding{55}}
\centering
\caption{Quantitative comparisons on CLIP~\cite{radford2021learning} similarity
with other methods.}
\vspace{0pt}
\setlength{\tabcolsep}{12pt}
    \begin{tabular}{l|ccc}
    \toprule
    Methods & ViT-L/14 $\uparrow$ & ViT-bigG-14 $\uparrow$ &Generation Time $\downarrow$ \\
    \midrule
    Shap-E~\cite{jun2023shap} & 20.51 & 32.21& 6 seconds\\
    DreamFusion~\cite{poole2022dreamfusion} & 23.60& 37.46 &  1.5 hours \\
    ProlificDreamer~\cite{wang2023prolificdreamer} & 27.39&42.98  & 10 hours\\
    Instant3D~\cite{li2023instant3d} & 26.87 &41.77  & 20 seconds\\
    Ours & 27.23 $\pm$ 0.06 & 41.88 $\pm$ 0.04& 15 minutes  \\
    \bottomrule
    \end{tabular}
\label{tab: abla-iters}
\end{table*}

\begin{figure*}[t!]
   \centering
   \includegraphics[width= \linewidth]{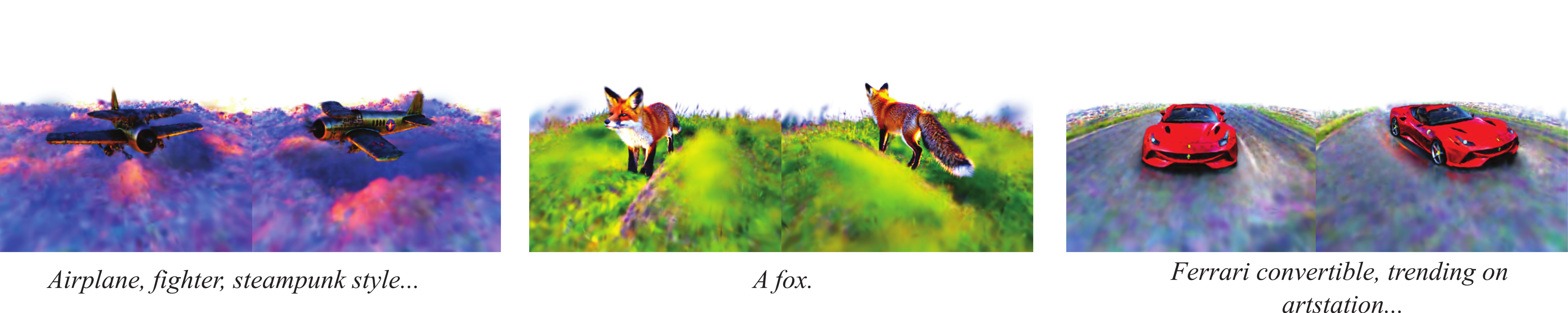}
   \caption{Results of generation with ground.}
   \label{fig: ablabg}
\end{figure*}

\begin{figure*}[t!]
   \centering
   \includegraphics[width= \linewidth]{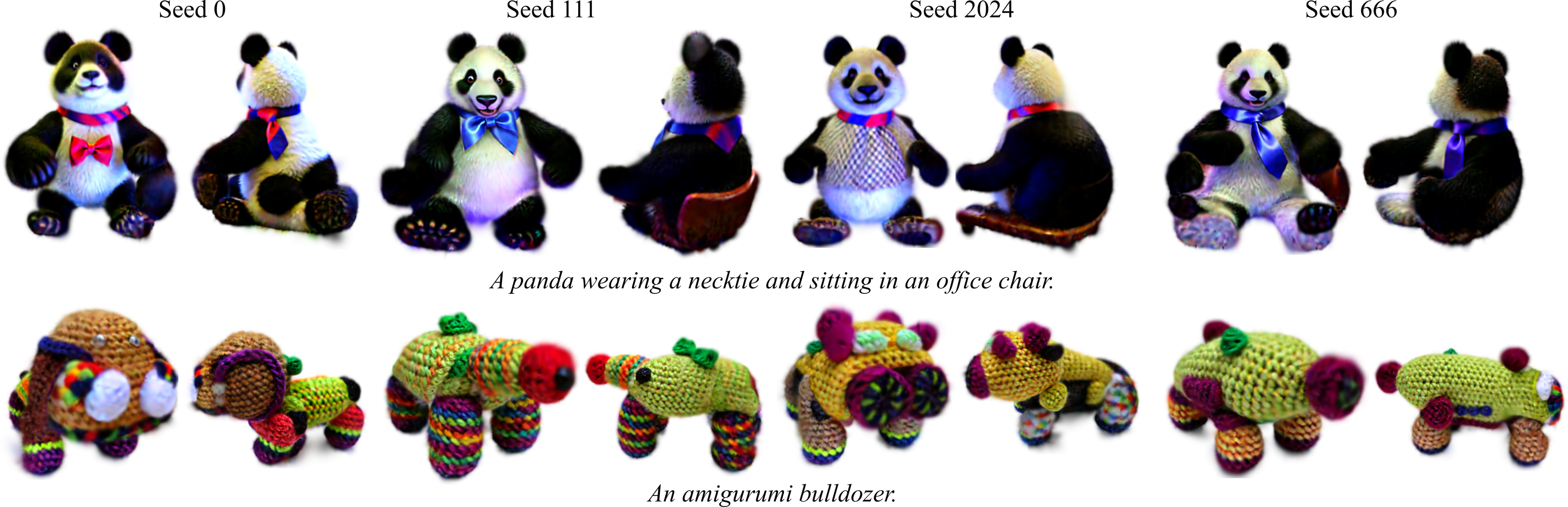}
   \caption{Results of the diversity of our method.}
   \label{fig: ablaseed}
\end{figure*}

\begin{figure}[t!]
   \centering
   \includegraphics[width= \linewidth]{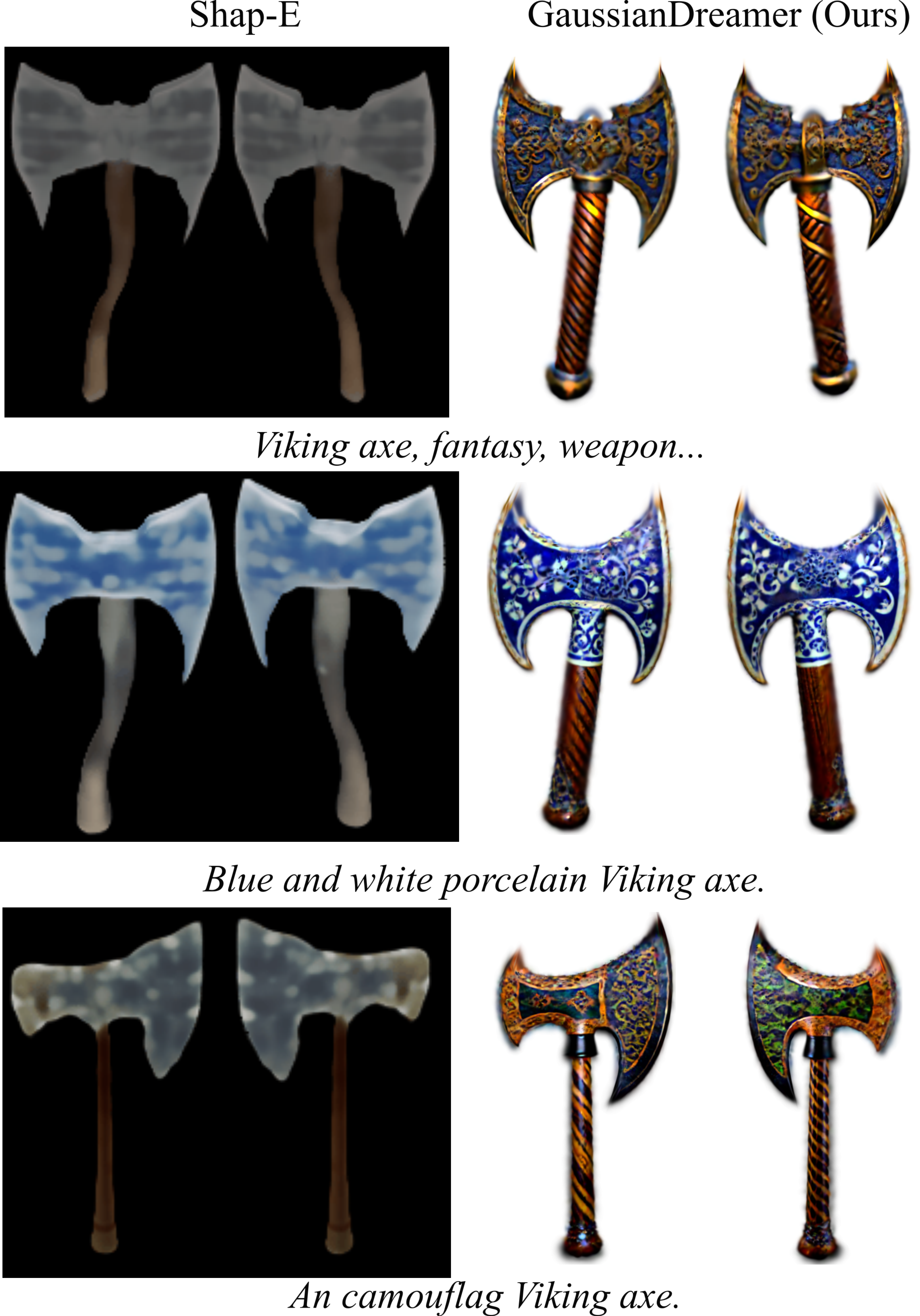}
   \caption{Results of generation with more fine-grained prompts.}
   \label{fig: abladomain}
\end{figure}

\begin{figure}[t!]
   \centering
   \includegraphics[width= \linewidth]{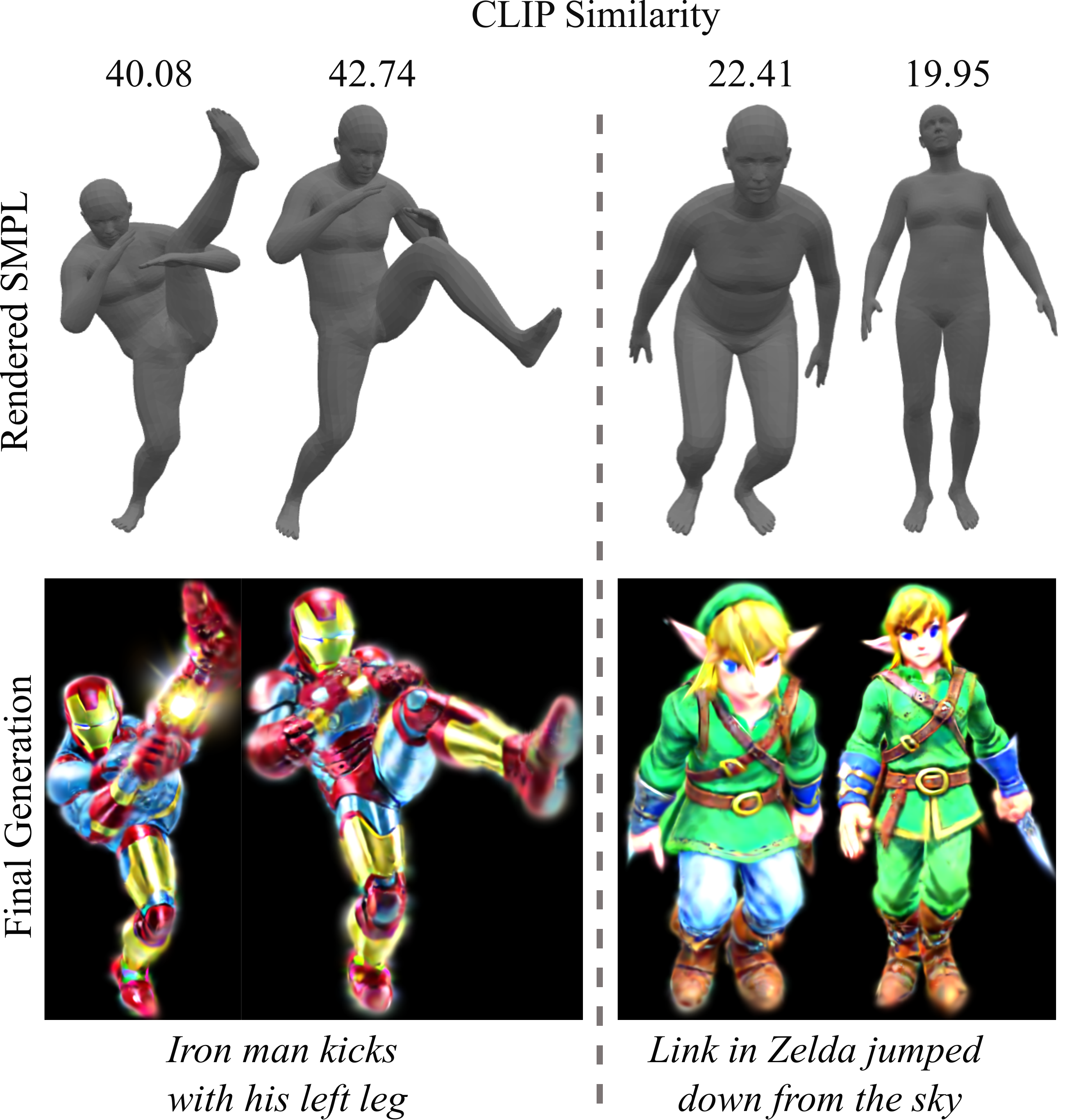}
   \caption{Avatar generation.}
    \label{fig: clip_smpl}
\end{figure}

\begin{figure}[t!]
   \centering
   \includegraphics[width= \linewidth]{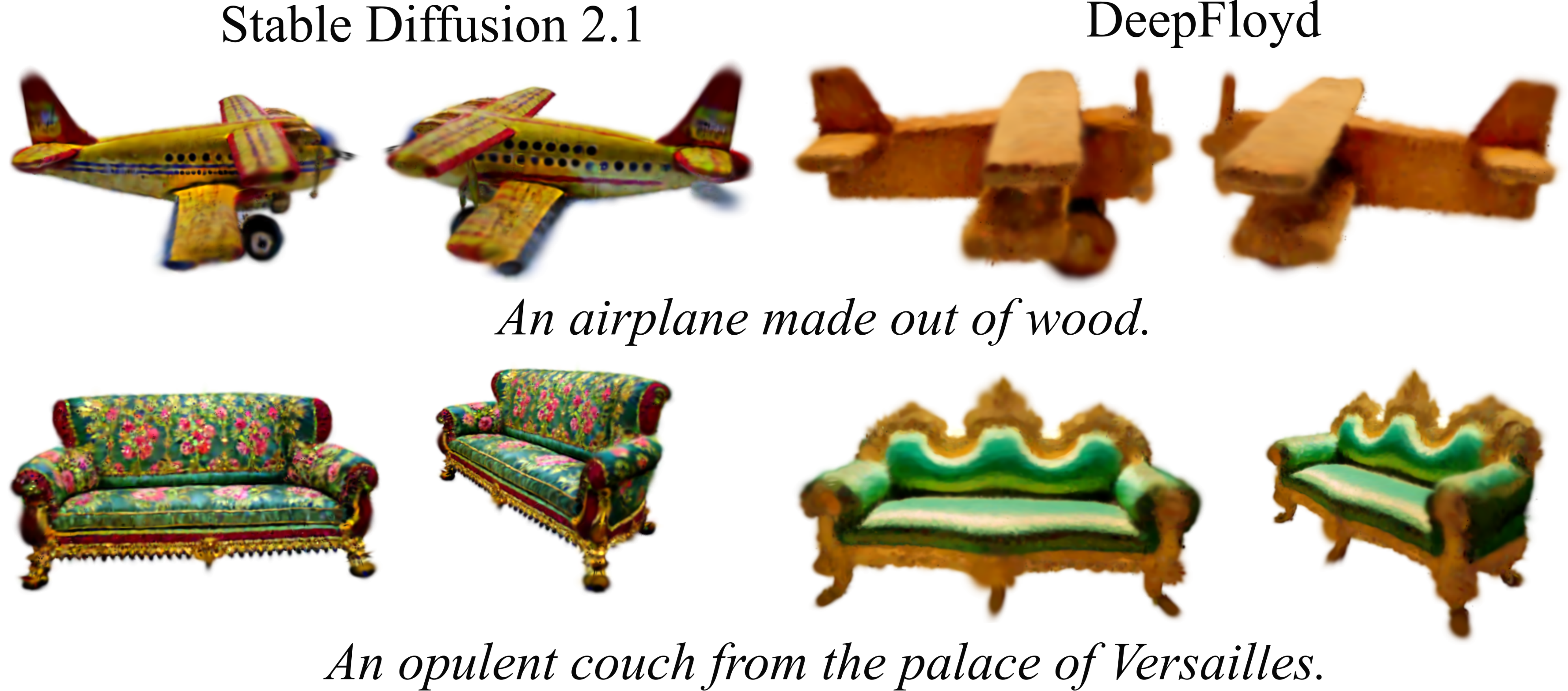}
   \caption{Ablation studies of optimizing 3D Gaussians with different 2D diffusion models.}
   \label{fig: ablaif}
    \vspace{-12pt}
\end{figure}

\begin{figure}[t!]
   \centering
   \includegraphics[width= 0.8\linewidth]{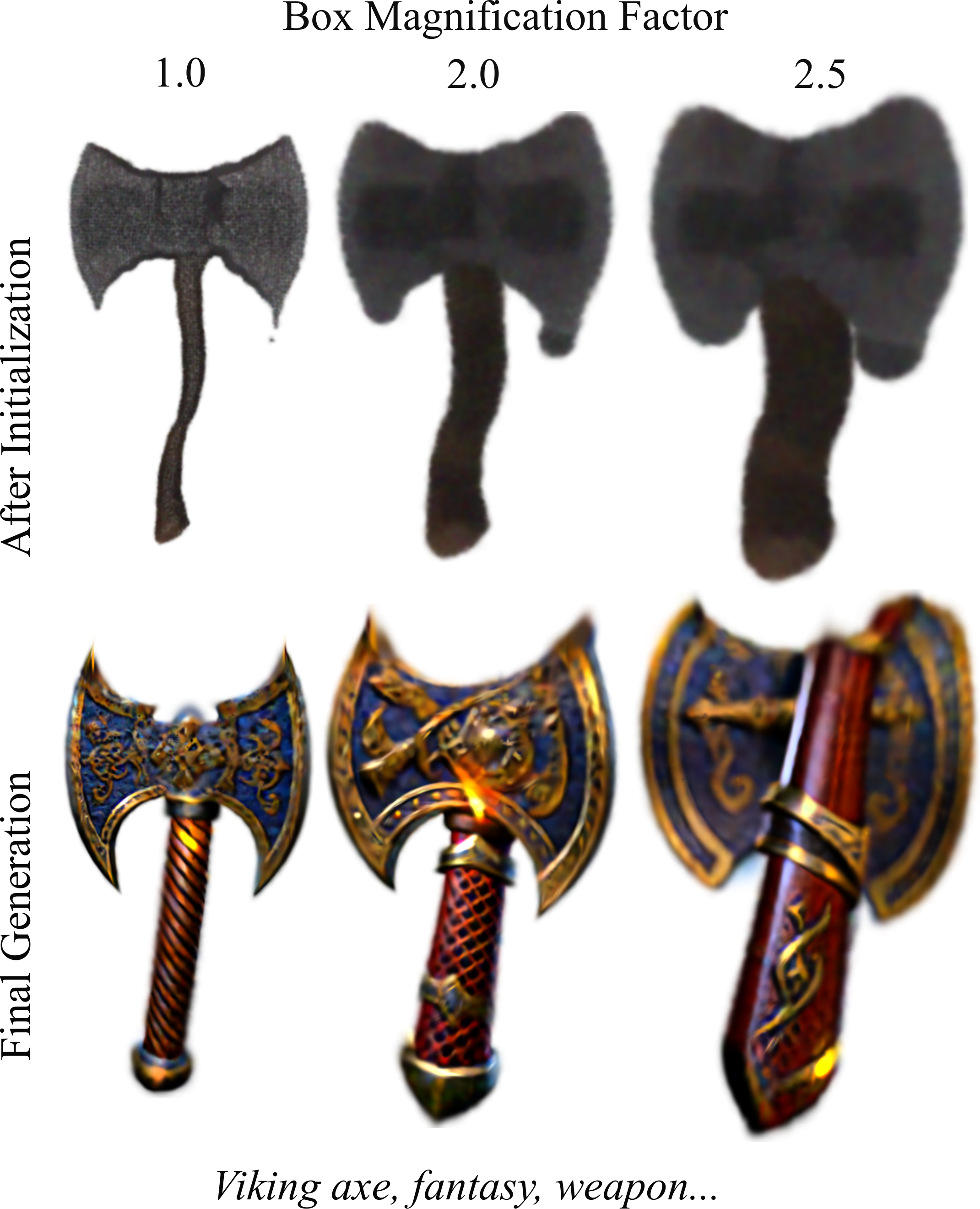}
   \caption{Ablation on the size of the box.}
   \label{fig: box}
\end{figure}

\begin{figure}[t!]
   \centering
   \includegraphics[width= 0.9\linewidth]{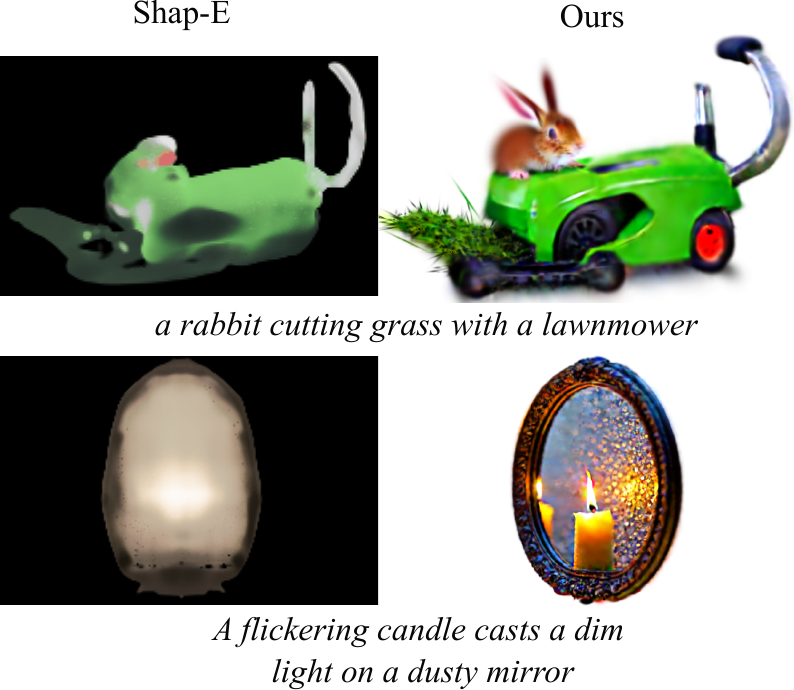}
   \caption{Generation with complex prompts.}
   \label{fig: more_com}
\end{figure}

\subsection{More Results}
\paragraph{Quantitative Comparisons.}
In Tab.~\ref{tab: abla-iters}, we use CLIP~\cite{radford2021learning} similarity to quantitatively evaluate our method. The results of other methods in the table come from the concurrent Instant3D~\cite{li2023instant3d} paper. The results of Shap-E~\cite{jun2023shap} come from the official source, while DreamFusion~\cite{poole2022dreamfusion} and ProlificDreamer~\cite{wang2023prolificdreamer} results come from implementation by threestudio~\cite{threestudio2023}. The implementation version of DreamFusion is shorter in time than the official report we mention in the main text. During the evaluation, we use a camera radius of 4, an elevation of 15 degrees, and select 120 evenly spaced azimuth angles from -180 to 180 degrees, resulting in 120 rendered images from different viewpoints. We follow the Instant3D settings, randomly selecting 10 from the 120 rendered images. We calculate the similarity between each selected image and the text and then compute the average for 10 selected images. It's worth noting that when other methods are evaluated, 400 out of DreamFusion's 415 prompts are selected. This is because some generations failed, so our method is disadvantaged during evaluation on all 415 prompts from DreamFusion. We use two models, \emph{ViT-L/14} from OpenAI~\cite{Radford2021LearningTV}~\footnote{https://huggingface.co/openai/clip-vit-large-patch14} and \emph{ViT-bigG-14} from OpenCLIP~\cite{schuhmann2022laionb,ilharco_gabriel_2021_5143773}~\footnote{https://github.com/mlfoundations/open\underline{ }clip}, to calculate CLIP similarity. Our method is superior to all methods except ProlificDreamer, but it is 40 times faster than ProlificDreamer in generation speed.
As shown in Fig~\ref{fig: i3d}, our method shows notably better quality and details than a concurrent work Instant3D but the CLIP similarity increases marginally.

\paragraph{Generation with Ground.}
When initializing, we add a layer of point clouds representing the ground at the bottom of the generated point clouds. The color of the ground is randomly initialized. Then, we use the point clouds with the added ground to initialize the 3D Gaussians. Fig.~\ref{fig: ablabg} shows the results of the final 3D Gaussian Splatting~\cite{kerbl3Dgaussians}.

\paragraph{Diversity.}
In Fig.~\ref{fig: ablaseed}, we demonstrate the diversity of our method in generating 3D assets by using different random seeds for the same prompt.

\paragraph{Generation with More Fine-grained Prompts.}
More refined prompts are used to generate 3D assets, as shown in Fig.~\ref{fig: abladomain}. It can be seen that Shap-E~\cite{jun2023shap} generates similar results when given different descriptions of the word "axe" in the prompt. However, our method produces 3D assets that better match the prompt.

\paragraph{Automatically Select A Human Model.}
As shown in Fig~\ref{fig: clip_smpl}, we attempt to use CLIP to guide the selection of the initialized human body model, by computing the similarities between images rendered from the generated SMPL models and the text prompt. We can achieve good rendering effects on various human body models. It would also be a promising direction to extend the assets to dynamic ones with the sequence of generated human body models.

\subsection{More Ablation Studies}
\paragraph{2D Diffusion Model}
During the process of optimizing 3D Gaussians with a 2D diffusion model, we perform ablation on the 2D diffusion models we use, specifically \emph{stabilityai/stable-diffusion-2-1-base}~\cite{rombach2022high}~\footnote{https://huggingface.co/stabilityai/stable-diffusion-2-1-base} and \emph{DeepFloyd/IF-I-XL-v1.0}~\footnote{https://huggingface.co/DeepFloyd/IF-I-XL-v1.0}. Fig.~\ref{fig: ablaif} shows the results of the ablation experiment, where it can be seen that the 3D assets generated using the \emph{stabilityai/stable-diffusion-2-1-base} have richer details.

\paragraph{Box Size in Point Growth}
In Fig~\ref{fig: box}, we conduct an ablation experiment on the box size, where a larger box leads to a fatter asset along with a more blurry appearance.

\subsection{More Discussions}
\paragraph{Limitations Introduced by the 3D Datasets.}
Fig~\ref{fig: more_com} shows the generation results of complex prompts. The domain-limited 3D diffusion model can only generate parts of the desired object with rough appearances. Our method completes the remaining part and provides finer details by bridging the domain-abundant 2D diffusion model.

\paragraph{Recent Works.}
We discuss with more related work. Our focus is to connect the 3D and 2D diffusion models, fusing the data capacity from both types of diffusion models and generating 3DGS-based assets directly from text. DreamGaussian~\cite{tang2023dreamgaussian} finally generates mesh-based 3D assets from an image or an image generated from text, which can be orthogonal to our method. There is a possibility of a combination in the future. NerfDiff~\cite{gu2023nerfdiff} uses a 3D-aware conditional diffusion to enhance details. DiffRF~\cite{muller2023diffrf} employs 3D-Unet to operate directly on the radiation field, achieving truthful 3D geometry and image synthesis. 3DDesigner~\cite{li20223ddesigner} proposes a two-stream asynchronous diffusion module, which can improve 3D consistency.


\end{document}